\pdfoutput=1
\documentclass[11pt]{article}

\usepackage{acl2023}
\usepackage{times}
\usepackage{latexsym}
\usepackage{graphicx}
\usepackage{amssymb}
\usepackage{comment}
\usepackage{booktabs}

\usepackage[T1]{fontenc}
\usepackage[utf8]{inputenc}

\usepackage{microtype}
\usepackage{inconsolata}

\usepackage{pifont}  % https://ftp.kddilabs.jp/CTAN/macros/latex/required/psnfss/psnfss2e.pdf

\usepackage{amsthm}

\theoremstyle{definition}
\newtheorem{formulation}{Formulation}[section]

\title{On Degrees of Freedom in Defining and Testing \\ Natural Language Understanding} %: Response-Dependent Interpretation and Validity Argument}

\author{Saku Sugawara \\
  National Institute of Informatics\\ %, Tokyo, Japan \\
  \texttt{saku@nii.ac.jp} \\\And
  Shun Tsugita \\
  Nagoya University \\ %, Nagoya, Japan \\
  \texttt{tsugita.shun.u7@f.mail.nagoya-u.ac.jp} \\}

\begin{document}
\maketitle
\begin{abstract}
Natural language understanding (NLU) studies often exaggerate or underestimate the capabilities of systems, thereby limiting the reproducibility of their findings.
These erroneous evaluations can be attributed to the difficulty of defining and testing NLU adequately.
In this position paper, we reconsider this challenge by identifying two types of researcher degrees of freedom.
We revisit Turing's original interpretation of the Turing test and indicate that an NLU test does not provide an operational definition; 
it merely provides inductive evidence that the test subject understands the language sufficiently well to meet stakeholder objectives.
In other words, stakeholders are free to arbitrarily define NLU through their objectives.
To use the test results as inductive evidence, stakeholders must carefully assess if the interpretation of test scores is valid or not.
However, designing and using NLU tests involve other degrees of freedom, such as specifying target skills and defining evaluation metrics.
As a result, achieving consensus among stakeholders becomes difficult.
To resolve this issue, we propose a validity argument, which is a framework comprising a series of validation criteria across test components.
By demonstrating that current practices in NLU studies can be associated with those criteria and organizing them into a comprehensive checklist, we prove that the validity argument can serve as a coherent guideline for designing credible test sets and facilitating scientific communication.
\end{abstract}

\section{Introduction}
\label{sec:intro}

Large-scale pretrained language models, also known as foundation models \cite{bommasani-etal-2021-opportunities}, have advanced significantly, leading to systems that are performing increasingly well at various natural language understanding (NLU) tasks and offering real-world applications \cite{devlin-etal-2019-bert, brown2020language,ouyang-etal-2022-training}.
However, these systems are often erroneously claimed to have human-level understanding \cite{bender-koller-2020-climbing}.
In other cases, their failures in specific situations are presented as systemic inadequacy, while ignoring their excellent performances in certain tasks \cite{jia-liang-2017-adversarial,bowman-2022-dangers}.

Such exaggerations may stem from unjustified assumptions about the capability of NLU systems.
Accordingly, researchers have attempted to improve the benchmarking of NLU through discussions to define language understanding \cite{bender-koller-2020-climbing,bisk-etal-2020-experience,merrill-etal-2021-provable} and practices, such as using auxiliary tasks for sanity check \cite{ribeiro-etal-2020-beyond}, decision boundary evaluation \cite{gardner-etal-2020-evaluating}, and dataset sourcing \cite{bowman-dahl-2021-will,kiela-etal-2021-dynabench}.
Such efforts have motivated researchers to reconsider and revise the concept and scientific study of NLU \cite{lakatos1976}.
Nevertheless, we need a more comprehensive guideline for better scientific communication.

In this study, we rethink this challenge in terms of researcher degrees of freedom, aiming to reframe the definition and evaluation of NLU and provide a pathway to better benchmarking. % researchers to improve their NLU studies.
We begin by revisiting a recent discussion by \citet{bender-koller-2020-climbing}, who define language understanding as the link between linguistic form (i.e., symbolic information) and the communicative intent of the speaker. 
They support their argument by presenting a thought experiment called the Octopus test, which is designed to resemble the Turing test.
In the Octopus test, an intelligent deep-sea octopus that can only use symbolic information (an analogy of language models trained only on textual corpora) tries to mimic a person in conversations.
The objective of the test is to ascertain whether the octopus can deceive the other person even when that person is defending themselves against an angry bear.
\citet{bender-koller-2020-climbing} argue that the octopus cannot achieve language understanding because it cannot pass the test in that situation (as a proof of existence).
% Current language models trained only on Internet corpora cannot learn the meaning from language form alone.
However, we find that the Octopus test may not be a valid thought experiment owing to initial concerns about the Turing test itself (e.g., even humans may sometimes fail the test).
Although we agree that symbol grounding plays an important role, several other features contribute to language understanding, which takes various forms \cite{sahlgren-2021-singleton}.

Here, we see that the degrees of freedom in defining NLU are characterized by the \textit{response-dependent interpretation} of the Turing test \cite{proudfoot2020rethinking}.
Thus, interpreting the subject's behavior depends on how it appears to a counterpart interrogator in the test, i.e., we cannot use success in the test as a definite criterion of language understanding, which makes it challenging to construct an NLU test with an operational definition.
Instead, using a test merely enables us to obtain inductive evidence to achieve consensus among stakeholders.
We elaborate on this observation by revisiting the Turing and Octopus tests in Sections \ref{sec:turing-test} and \ref{sec:octopus-test}, respectively.
Then, we restate the testing of NLU in Section~\ref{sec:reframing} considering that its definition is inevitably arbitrary to observers.
In short, a test has to assess distinguishable behaviors in a specific domain rather than general NLU to make it easier for stakeholders to arrive at a consensus about the interpretation of the test results.

Although we can decide what capability we assess in a test, other degrees of freedom are present while designing the test and interpreting its results.
This problem can be understood in terms of psychological studies, in which choices arbitrarily taken by researchers heighten the chances of false positive results \cite{simmons2011false,wicherts2016degrees}.
Several previously proposed NLU practices facilitate detailed evaluation to lessen unjustified claims about NLU systems. However, they are not sufficiently well-organized to be deployed on a universal framework.
Therefore, we often choose convenient practices depending on specific situations. Consequently, the degrees of freedom that researchers have in evaluating NLU increase, thereby sacrificing the reproducibility of research \cite{munafo2017manifesto}.
In Section~\ref{sec:validity-argument}, we introduce the validity argument, which is a framework used in psychological and educational tests \cite{kane2006validation,chapelle2008test,cook2015contemporary}, to set out a guideline for designing, conducting, and using NLU tests.
We demonstrate that current practices are associated with inference steps in the validity argument and show that it may serve as a comprehensive, coherent guideline for developing or identifying actionable and beneficial practices to construct a better NLU test and use it properly. % providing a comprehensive and universal roadmap for constructing better NLU tests.

Our major contributions are two-fold:
\begin{itemize}
    % \item Using the response-dependent interpretation of the Turing test, we rethink the NLU evaluation. According to the revised concept, a test need not provide a concrete definition of NLU; it is considered effective if it presents inductive evidence, using which stakeholders can concur with the interpretation of the target behavior.
    \item Using the response-dependent interpretation of the Turing Test, we rethink the NLU evaluation. In our revised formulation, a test does not provide a concrete definition of NLU but presents inductive evidence, using which stakeholders can concur with the interpretation of the target behavior.
    \item As a tool for designing and using NLU tests, we introduce the validity argument with a checklist of 16 questions, which helps stakeholders to collect and interpret validity evidence coherently, thereby encouraging more reproducible research. %and scientific communication.  % and use
\end{itemize}

\section{Turing Test Revisited}
\label{sec:turing-test}

% In this section, we present an initial description of the Turing Test and review the previous research on it.
% Thereafter, outline \citet{proudfoot2020rethinking}'s interpretation of the Turing Test, which we believe is more promising than the standard behaviorist interpretation.
% Finally, we discuss the Turing Test's problems that we consider crucial.

\subsection{Imitation Game}

\citet{turing1950computing} proposes a game known as the Imitation Game, which is a conversational test to examine a machine's intelligent behavior: % (see Appendix~\ref{app:turing-test-definition} for its definition).
\begin{quote}
A human questioner speaks in natural language to one machine with another human for a certain period.
These participants are isolated and can communicate only in text through the display.
The topics of conversation and length of questions are unrestricted.
The human and machine respond to the interrogator's questions in a way that makes them appear human.
The interrogator wins if he identifies the machine as a machine and the human as a human.
\end{quote}
\noindent According to the Turing test, if a machine wins the Imitation Game more or less reliably, it passes the test, indicating that the machine can be considered intelligent.
%
% Before going into the details,
% Furthermore,
% Note that the test employs a sufficient condition for intelligence, not a necessary condition.
% Generally, the odds against machines are enormous.

Machines may fail the Turing test owing to factors irrelevant to intelligence.
For example, a human participant may perform poorly in pretending to be a machine. Similarly, machines may exhibit a process that can be described as thinking but differs significantly from the process performed by humans.
To circumvent these problems, the Turing test focuses solely on a sufficient condition for intelligence \cite[][Copeland 2004, p. 442]{turing1950computing}.

\subsection{Response-Dependent Interpretation}

To evaluate the Turing test, we must clarify the interpretation of intelligence that argues that thinking is performed intelligently if the Imitation Game is played well.
However, Turing's interpretation of intelligence is controversial.
In the following section, we discuss two interpretations: the standard behaviorist interpretation and the more recent response-dependent interpretation.

According to the standard interpretation, the Turing test is a behaviorist test of intelligent thinking.
In general, if a machine behaves as though it is intelligent, it is intelligent \cite{block1981psychologism}. % Block states this concept \cite{block1981psychologism}.
%
% \begin{quote}
%     Intelligence (or more accurately, conversational intelligence) is the disposition to produce a sensible sequence of verbal responses to a sequence of verbal stimuli, whatever they may be. \cite[][]{block1981psychologism}  % \cite[][p.11]{}
% \end{quote}
%
% Behaviorism was undoubtedly popular in the 1940s and 50s.
The Turing test is reminiscent of behaviorism, as it requires evidence for intelligent thinking to be a publicly observable behavior.
However, at least two reasons exist to contradict the standard interpretation.
First, Turing does not suggest that the mental vocabulary of ``intelligence'' or ``thinking intelligently'' is definable based on observable or behavioral terms \cite{davidson1990turing}.
Second, and more importantly, the game tests the interrogator's response and not the machine's behavior.
In other words, the Turing test does not examine whether a machine can perform a specific cognitive task but how effectively the interrogator is fooled \cite{proudfoot2020rethinking}.
% Combining the above two reasons, behaviorism is not expected to be the key motivation for the idiosyncratic design of the Turing Test.

\citet{proudfoot2020rethinking} provides an alternative interpretation of Turing's view of intelligence, which is considered more promising than the standard behaviorist interpretation.
Proudfoot's exegesis begins with the observation that the first version of the Imitation Game appeared in the final section, ``Intelligence as an emotional concept'' of \citet{turing1948intelligent}. 
In this text, Turing states: % the following:
% 
% \begin{quote}
\textit{The extent to which we regard something as behaving in an intelligent manner is determined as much by our own state of mind and training as by the properties of the object under consideration} \cite[][Copeland 2004, p. 431]{turing1948intelligent}.
%\end{quote}
%
\noindent Using the phrase ``intelligence as an emotional concept,'' Proudfoot asserts that intelligence is a response-dependent property using current philosophical terminology. Response-dependent properties are those that depend on human responses under certain specified conditions.
Secondary qualities, such as color, and values, such as beauty, are examples of response-dependent properties. 

A simple response-dependence theory of color may be stated as follows:

\begin{quote}
$x$ is red if and only if, in normal conditions, $x$ looks red to normal subjects.
\end{quote}

\noindent Applying this formulation to intelligence leads to a simple response-dependence theory of intelligence: 

\begin{quote}
$x$ is intelligent if and only if, in normal conditions, $x$ appears intelligent to normal subjects.
\end{quote}

\noindent Identifying the ``normal'' conditions is a typical problem faced by response-dependent theories.
However, the setting of the Imitation Game appears to naturally reflect the normal conditions associated with intelligence.
Selecting a ``normal'' interrogator is also challenging from the perspective of response-dependent theories.
Therefore, Turing chose average citizens to play the role of interrogator.
Thus, Proudfoot suggests that a response-dependent approach that is suited to the Turing test appears as follows:

\begin{formulation}\label{form2-1}
    $x$ is intelligent if, in an unrestricted computer-imitates-human game, $x$ appears intelligent to an average interrogator.
\end{formulation}

% Proudfoot technically provides a subtler interpretation than that indicated by the above discussion. However, 
\noindent We assume that Formulation \ref{form2-1} captures the view of intelligence that underlies the Turing test.

\subsection{Crucial Problems with the Turing Test}
\label{sec:turing-test-problems}

We are inclined to consider intelligence as a response-dependent property, as suggested by Turing and Proudfoot.
% However, it remains debatable whether an unrestricted computer-imitates-human game is an appropriate means for testing intelligence. \citet{hayes1995turing} point out several problems with the Turing test:
However, whether an unrestricted computer-imitates-human game is an appropriate means for testing intelligence remains debatable. \citet{hayes1995turing} point out several problems with the Turing test.

First, the test design is flawed because it is indeterminate what is being tested.
Even if a machine could reliably pass the test, we would not be able to ascertain whether the machine was truly intelligent, or the interrogator was not sufficiently clever to ask informative questions.
In short, the object of testing is unclear (Issue 1). % , and whether it can serve as a test of intelligence. 

Second, the Turing test indirectly identifies intelligence by distinguishing the participant as a human or machine.
This design forces machines to hide their inhuman abilities to impersonate humans.
Thus, the test focuses on identifying intelligent behavior that successfully deceives the interrogator.
% As Turing argued, lying and cheating may be signs of intelligence at first glance. However,
Examples of this type of behavior were observed in the Loebner competition, where the winner sometimes deliberately mistyped a word and subsequently backspaced it to correct it at a human typing speed purely to deceive its interrogator. % Alternatively, to behave as if it were a human with specific attributes, the system must implement characteristics that the human would know.
% It is counter-intuitive that such tricks are related to intelligence. 
Therefore, even if a machine could reliably pass the test, it might not be intelligent (Issue 2).

Finally, even humans cannot pass the Turing test under certain conditions.
Heuristics that are used to distinguish the behavior of machines from that of humans sometimes misguide interrogators.
For instance, judges in the Loebner competition identified a human as a machine because they produced extended, well-written paragraphs of informative text, which tends to be associated with inhuman skills in certain parts of our culture (Issue 3).
% Turing Test is not considered an inappropriate test for intelligence, not because it is easy to pass; conversely, passing the Turing Test may be extremely difficult. \citet{hayes1995turing} note that the difficulty in passing the test does not mean that the test is appropriate for determining whether a person has intelligence. They conclude that a Turing Test is a test of the ability to pass the corresponding Turing Test, which is circular.

In summary, we agree with \citet{hayes1995turing}, who argue that passing the Turing test should not be the goal of AI research.
Moreover, passing the test is not a necessary condition for using it as a real-world technology for humans or psychologically investigating human intelligence.

\section{Rethinking the Octopus Test}
\label{sec:octopus-test}

\subsection{Octopus Test}

% The evaluation of language understanding is a long-standing goal in the fields of AI and NLP.
% Similar to the shortcomings of Turing’s interpretation of intelligence, definitions of language understanding in the fields of philosophy and linguistics are also considered inadequate.
Similar to the difficulty of defining intelligence in the Turing Test, the definition of language understanding has also been difficult in the fields of philosophy and linguistics.
% It is quite difficult to provide a rigid and satisfactory formulation.
\citet{bender-koller-2020-climbing} indicate that current hypes relating to natural language processing (NLP) systems are partly because of this confusing and challenging concept of language understanding.
% They provide the following formulation regarding language understanding: 
They define meaning $M$ as $M \subseteq E \times I$, where $E$ is a set of possible forms and $I$ is a set of possible communicative intents, pragmatically constructed in addition to the conventional meaning. % provided by the semantics.

According to this definition, understanding the forms is inevitably accompanied by associating them with their communicative intents; therefore, \citet{bender-koller-2020-climbing} propose that systems that deal with the forms alone do not understand meaning by definition.
They present a thought experiment known as the Octopus test to explain this argument: % (see Appendix~\ref{app:octopus-test-definition} for its definition).  %, which is summarized as follows:

\begin{quote}
Suppose speakers A and B drift ashore on two separate uninhabited islands.
There is a communication device on each island connected by a submarine cable that enables A and B to communicate.
At the bottom of the sea, there is an octopus.
Although this octopus does not understand their language, by intercepting the cable communication, it finds statistical patterns from their conversation and learns to predict how B would answer A accurately after a certain period.
At some point, the octopus cuts the cable and tries to respond to A while pretending to be B.
Will the octopus be able to continue to respond to A without raising suspicions?
\end{quote}

\noindent In this test, \citet{bender-koller-2020-climbing} use the deep-sea octopus as an analogy for pretrained language models that are trained only on textual corpora.
The octopus does not have access to sensory data associated with the speaker's communicative intents, which are essential in the authors' definition of meaning (i.e., the link between forms and communicative intents).

\citet{bender-koller-2020-climbing} argue that, in certain situations, the octopus might be incapable of responding to A without arousing suspicion.
For example, if A asks how one can build a coconut catapult or what to do if a bear appears, the octopus can offer a convincing answer only if it accurately understands A's situation.
However, because the octopus does not have the means to deal with novel information and unforeseen events beyond text, it cannot provide a sufficient answer to questions beyond the scope of what it has learned.
Hence, A possibly determines that B is not human.\footnote{See also Appendix~\ref{app:symbol-grounding} for our brief discussion on symbol grounding.}
% \citet{bender-koller-2020-climbing} conclude that it is impossible to achieve language understanding using the statistical language models.

However, should we use the Octopus test as a test of language understanding in a similar capacity to the original Turing test?
Although the Octopus test does not place specific conditions on the speakers A and B, we attempt to reframe the definition of language understanding implied by \citet{bender-koller-2020-climbing}, according to Formulation~\ref{form2-1}:

\begin{formulation}\label{form3-2}
    $x$ understands language if, in an unrestricted imitation game, $x$ appears to understand language to an average interrogator.
\end{formulation}

\noindent Nevertheless, even if the Octopus test conforms to this formulation, it has severe drawbacks similar to those of the original Turing test.

\subsection{Issues in Octopus Test}
\label{sec:octopus-test-issues}

% The Turing test is designed to determine whether a subject thinks intelligently. In contrast, the Octopus test checks whether it understands the language. Nevertheless, 
The Octopus test checks whether it understands the language, but Issues 1 to 3 of the Turing test in Section~\ref{sec:turing-test-problems} also apply to the Octopus test.
First, the ability of this test form to evaluate intelligent behavior is questionable, as the inability to converse does not mean that the subject is not intelligent or that the subject cannot understand language.
This argument aligns with the singleton fallacy \cite{sahlgren-2021-singleton}: the ability to understand language takes various forms.
% For example, 
Pretrained language models may be able to exhibit behavior that can be regarded as language understanding by average interrogators, and no clear evidence for denying such an interpretation is available (Issue 1). % (\textbf{Issue 3.1}).  % definitive

Second, both tests in our formulation define the ability to deceive the average interrogator as a sufficient condition for being intelligent or understanding language; however, no necessary condition has been formulated.
Hence, by simply observing that the subject fails in the imitation of humans, we cannot conclude that it does not exhibit NLU because a subject may understand the language even if it does not pass the Octopus test (Issue 2). % (\textbf{Issue 3.2}).

Finally, we cannot prove that a situation exists where the octopus fails to impersonate human B.
As \citet{sahlgren-2021-singleton} argue, in the situation with a bear, the pretrained language model can possibly learn web traffic and generate a meaningful answer to the question of how to use a stick to protect oneself from the bear.
Similarly, the octopus may be able to meaningfully communicate with A in any situation.
Even if we accept the possibility of the octopus's failure, we cannot use this test owing to Issue 3 in Section~\ref{sec:turing-test-problems}, that is, even humans may fail the test. % the deception-style test cannot ascertain the presence of NLU.

\section{Reframing the Response-Dependent Interpretation of NLU Tests}
\label{sec:reframing}

In this section, we investigate how NLU should be practically tested under the response-dependent interpretation of the Turing test.
Considering the three issues outlined in Section~\ref{sec:octopus-test-issues}, our goal is to formulate a necessary and sufficient condition of language understanding that does not contradict the actions performed in an NLU study.

First, to achieve our objective, we summarize the corresponding requirements of the formulation:
\begin{enumerate}
    \item Specify what type of language understanding behavior is to be tested. This specification may include target tasks, skills, domains, and data format (for Issue 1).
    \item Impose behavioral tests as a means to evaluate NLU, which provides evaluation metrics that are objective to the observers (for Issue 2).
    \item Consider the response-dependent interpretation of the behavioral test to avoid directly defining specific linguistic behavior as language understanding (for Issue 3).
\end{enumerate}

\noindent Based on these requirements, we reformulate and elucidate the response-dependent interpretation of language understanding:

\begin{formulation}\label{form4-3}
    $x$ understands language under the condition $c$ if and only if the subjective probability of an average observer for the hypothesis, $x$ understands language under $c$, is higher than a threshold, where the hypothesis is supported by evidence obtained by the performance of $x$ on a test under $c$.
\end{formulation}

\noindent According to the three requirements, we elaborate on this formulation as follows.

\paragraph{1. Specifying Target Linguistic Behavior}
\citet{raji-etal-2021-ai} suggest that tests evaluating general language understanding might be impractical owing to their propensity to make false claims.
Humans generally identify language understanding in the various behaviors of others, and the success of that behavior is determined depending on stakeholder objectives.
% Furthermore, even humans do not have a practical, versatile test for general language understanding designed for humans or other agents.
Owing to this broad scope, designing a practical, versatile test may be unrealistic.
Therefore, we argue that NLU should be decomposed into distinguishable capabilities by specifying a target condition $c$, including skills, tasks, data sources, input and output format, and potential applications.

\paragraph{2. Using Behavioral Tests}
To address the issues of the Turing test, \citet{levesque2014behaviour} argues that a behavioral test should be used to test the commonsense reasoning of machines.
% Similarly, we intend to use a behavioral test to evaluate NLU because we cannot deny that it provides objective measures (i.e., \textit{evidence obtained by the performance of $x$ on a test}).
Similarly, we can use a behavioral test to evaluate NLU because it provides objective measures (i.e., \textit{evidence obtained by the performance of $x$ on a test}).
However, we do not deny the possibility of using a test that inspects a machine's internal properties to effectively evaluate NLU (e.g., using interpretation methods for deep learning models), although such a test would have to define objective criteria for target internal properties to ensure stakeholder consensus.

\paragraph{3. Response-Dependent Interpretation}
This interpretation does not provide an explicit definition of the expected target behavior but requires an average observer to observe the subject's output in a test.
The statement \textit{the subjective probability of an average observer [...] is higher than a threshold} indicates that stakeholders or their representative experts agree on their interpretation of test results (i.e., the observed behavior is successful). % the single average human of ordinary people but is rather conceptual.
In other words, (the approximation of) the average speaker obtains certain information regarding the subject's behavior $x$ and calibrates the subjective probability for the hypothesis.
In this formulation, a linguistic test for language understanding does not provide an operational definition; instead, it provides inductive evidence that the speaker can use for their calibration \cite{moor1976analysis,moor2001status}.

% Here the domain determines the boundary of possible linguistic behavior by specifying the modality and content of the input and output. % , including the vocabulary and contextual information.
% In NLP, this corresponds to specifications such as a source of data, text format, and tasks.
% According to the definition of a domain, an average speaker can view language understanding as a specific type of behavior, making it easier to determine whether or not the observed behavior is successful.
% Such tests no longer need conversational or deceptive styles.  % to exhibit

% In terms of the test that should be created,
% When we specify the condition $c$, this does not necessarily yield a useful test for an average speaker to interpret the result and calibrate the subjective probability.
% If the test consists of ambiguous or complex instances, the average speaker will be uncertain of the interpretation of the target behavior.
% Therefore, a test should consist of instances of moderate difficulty (i.e., not too easy or difficult) for subject $x$ that are unambiguous to humans.
% A test designed under these criteria can be seen as a \textit{benchmark}.

% \paragraph{Philosophical Implication}
Our formulation may be seen as a meta-definition of NLU in the sense that the stakeholders are required to determine their own definition of language understanding in line with their objectives.
In other words, it provides an epistemological view of language understanding and avoids making a commitment to what language understanding is.\footnote{We follow the terminology of \citet[][Section 2.6.3]{bommasani-etal-2021-opportunities}.
Epistemology is related to \textit{how we know that an agent has achieved the relevant type of language understanding}, while metaphysics concerns \textit{what it would mean for an agent to achieve language understanding}.}
This flexibility in interpretation is possible because the stakeholders can define their own characterizations of language understanding and instantiate intended behavior into target tasks based on their objectives.\footnote{For example, if stakeholders adopt internalism, their test requires the subject to have consistent internal representations for intended tasks.
See also \citet[][Section 2.6.3]{bommasani-etal-2021-opportunities} for an overview of their metaphysical characterizations of language understanding that may fit the development and deployment of foundation models.}
However, this formulation permits several degrees of freedom in how stakeholders define what language understanding is, which can make our communication about NLU obscure.
Therefore, the designers and users of NLU tests have to precisely specify what behavior is evaluated in their tests, including the conditions $c$, and accordingly interpret the test results.
This effort improves the validity of resulting claims and thus contributes to avoiding over- and under-claiming.

\section{Validity Argument for Testing NLU}
\label{sec:validity-argument}

\begin{figure*}[t]
    \centering
    \includegraphics[width=\linewidth]{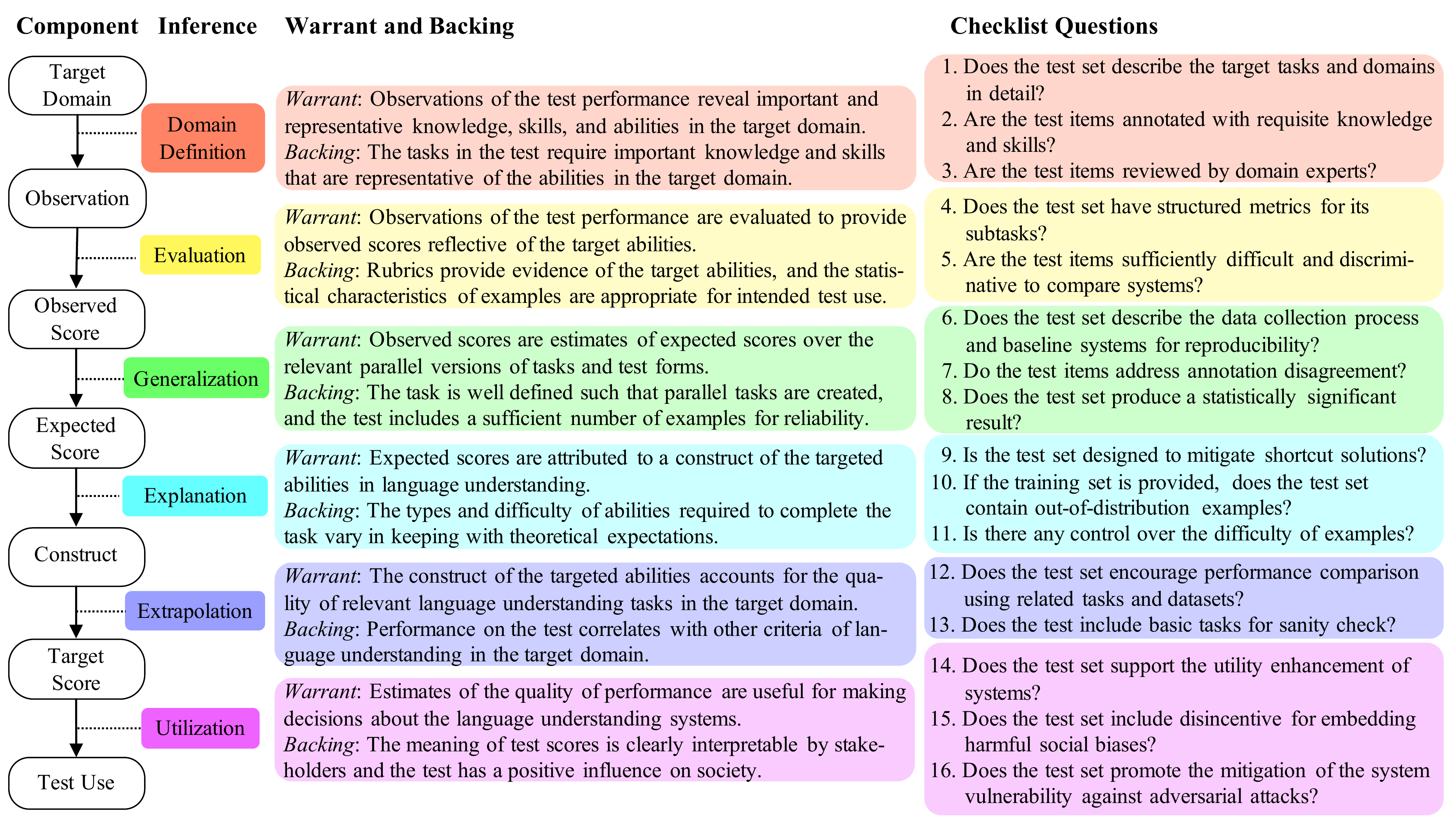}
    \caption{
        Overview of a validity argument for NLU, including inferences with its warrant and backing.
    }
    \label{fig:validity-argument}
\end{figure*}

In Formulation~\ref{form4-3}, the stakeholders employ certain measures to evaluate the behavior of a test subject and interpret those measures to arrive at a decision.
To support the interpretation of the measures (i.e., improve the stakeholders' confidence in the subject's target linguistic behavior), the test has to be well-designed to provide sufficient evidence from various perspectives.

Despite significant progress, NLU benchmarks often lack the ability to provide substantial evidence \cite{bowman-dahl-2021-will, raji-etal-2021-ai, dehghani-etal-2021-benchmark}.
To date, useful practices have been proposed to better connect observed scores and intended NLU behavior \cite[e.g.,][]{mccoy-etal-2019-right,gardner-etal-2020-evaluating}.
However, without guidelines on design and using a test, these practices cannot be organized coherently.
As a result, researchers can arbitrarily select practices most suited to their purpose while, intentionally or not, ignoring others.
This freedom of choice allows interpretations of test results that are not justified by reliable evidence.
Such unjustified interpretations are reminiscent of a problem in psychology known as researcher degrees of freedom \cite{simmons2011false,wicherts2016degrees}.
We need to tackle this challenge by developing a unified, comprehensive framework to overview necessary practices that validate the interpretation of test results.

% As we viewed in Formulation~\ref{form4-3}, NLU tests only provide inductive evidence rather than a definition of language understanding.
% For helping stakeholders to assess a machine's language understanding capability in a specific domain, NLU tests have to be designed to provide sufficient interpretation to exhibit validity for performing the target behavior. 

% However, the evaluation of language understanding on current NLU tasks is broken \cite{bowman-dahl-2021-will}; critical issues that have been pointed out prevent us from interpreting what capability is tested.
% Although many practices have been proposed so far to establish a better methodology, those methods are not well-organized yet.  % of NLU evaluation
% This leaves researchers ample room for freedom in choosing a practice that is most useful to their purpose and, intentionally or not, ignoring others.
% This allows the interpretation of test results that are not justified by reliable evidence.
% This reminds us of the problem known as researcher degrees of freedom in psychology.
% We need a unified, comprehensive framework to overview necessary practices that validate our interpretation of test results.

In psychological measurements, \citet{kane2006validation,kane-2012-validating} proposes the validity argument, which is a theoretical framework that guides the collection of evidence to validate the interpretation and use of test scores (Figure~\ref{fig:validity-argument}).
It decomposes the design, conduct, and use of a test into the seven components where transitions between them are performed as inferences supported by a warrant and its backing that follow Toulmin's formulation of arguments \cite{toulmin2003uses}.
The six inferences constitute the process of collecting and interpreting evidence that the test designers and users follow.
The validity argument has been employed in various fields, including linguistic tests \cite{chapelle2008test} and clinical exams \cite{cook2015contemporary}, but not in NLU.
See Appendix~\ref{app:validity-argument} for its background and terminology.

% Fortunately, we find that \textit{validity argument} (Figure~\ref{fig:validity-argument}) in psychological measurements deserves such a framework \cite{kane2006validation,kane-2012-validating,chapelle2008test}.  % educational
% Figure~\ref{fig:validity-argument} shows its overview. % of a validity argument.\footnote{Refer to \citet{chapelle2008test} for detailed formulation.}
% It decomposes the design, conduct, and use of a test into the seven components where a transition between them is performed as an inference supported by a warrant that a few assumptions underlie Toulmin's formulation \cite{toulmin2003uses}.  % components -> them, the formulation of 
% These inferences help stakeholders assess aspects of validity, leading to justifying their score interpretation and use.
% 
% In the following section, we overview the validity argument in terms of NLU, showing that current important practices can be associated with the inferences of the validity argument. Thus, the argument serves as a useful guideline to design and use NLU tests. %
In the following section, we apply the validity argument to the evaluation of NLU.
We associate current important practices with the inferences of the validity argument, show that the argument serves as a useful guideline to design and use NLU tests, and provide relevant checklist questions
Here, we take SQuAD \cite{rajpurkar-etal-2016-squad}, a question answering (QA) dataset consisting of crowdsourced questions written for Wikipedia articles to instantiate a validity argument and identify missing aspects.
Appendix~\ref{app:checklist-practices} elaborates on the checklist and relevant practices.

\paragraph{1. Domain Definition}
This inference requires performance observation in the test to reveal relevant knowledge and skills in the target domain, which contributes to providing a means to clarify the achievements and weaknesses of the test subjects at a theoretical level \cite{doshi-velez-kim-2017-rigorous,rogers-etal-2023-qa}. %\footnote{We use \textit{domain} in the same sense with that in Section~\ref{sec:reframing}.}
% This is supported at the theoretical level through analysis by experts who look at the test format and examples.
In SQuAD, the authors take 100 questions to annotate necessary reasoning types, including lexical and world knowledge, syntactic variation, and multiple-sentence reasoning.
However, given that reading comprehension involves other types of reasoning, such as temporal and causal \cite{dunietz-etal-2020-test}, this annotation may reveal that SQuAD can evaluate only a limited number of abilities in reading comprehension.
This observation implies that, ideally, the test examples need to be reviewed by domain experts (i.e., researchers of human reading comprehension) who may be able to indicate relevant knowledge and reasoning types that are missing from the dataset.
% The expert-in-the-loop design for collecting examples may be a promising direction to target intended linguistic phenomena~\cite{parrish-etal-2021-putting-linguist}, though we need to take care that the distribution of data is not unintentionally biased towards a limited variety of linguistic phenomena \cite{bowman-dahl-2021-will}. % that are well-known and -studied~\cite{bowman-dahl-2021-will}.

\paragraph{2. Evaluation}
This inference is related to the design of scores and characteristics of test examples.
SQuAD provides evaluation metrics that calculate word overlap between predicted and gold text spans. However, it only enables a one-dimensional interpretation of the system performance.
% In a similar reading comprehension task, HotpotQA \cite{yang-etal-2018-hotpotqa} offers a different metric for supporting facts, % a set of sentences necessary for predicting the correct answer, 
% which helps system developers test the intermediate step of reasoning in answering questions. % (i.e., finding necessary sentences) 
% The R4C dataset~\cite{inoue-etal-2020-r4c} extends this formulation by adding an entity-level prediction task that evaluates the reasoning process in detail.
Ideally, the task provides multiple metrics that correspond to target abilities or decomposed subtasks identified in the domain definition \cite[e.g.,][]{wolfson-etal-2020-break}.
This inference also requires the statistical characteristics of the test examples to be monitored.
% Depending on the test's intended use,
The test developers have to ensure that collected examples are discriminative and their differences are relevant to the target abilities.
Otherwise, stakeholders cannot associate the system performance difference with the target abilities.
% the examples have to be enough characteristics to discriminate systems to be tested in terms of abilities the test is intended to assess.
A promising way to meet this requirement is to use the item response theory, which models the difficulty and discriminability of examples using the responses of test subjects \cite{rodriguez-etal-2021-evaluation,vania-etal-2021-comparing}.
% Only several works use IRT for NLU tasks and provide practices for identifying discriminative and difficult examples \cite{lalor-etal-2016-building,rodriguez-etal-2021-evaluation,vania-etal-2021-comparing}.

\paragraph{3. Generalization}
This inference concerns the generalizability of the observed scores, which is characterized by their reproducibility and reliability. %\footnote{In this sense, the meaning of generalization here is different from that used for machine learning models.}
For reproducibility, the task specifications must be well-described to ensure that others can construct similar tasks for the same target.
Reporting specifications using a template \cite{gebru2021datasheets} would suffice to cover necessary facets such as the components, collection, and preprocessing of the dataset.
% If a paper proposing a new dataset includes baseline machine learning systems, ensuring the reproducibility of baseline systems would also be important \cite{pineau2021improving}.
For reliability, % a test needs a sufficient number of examples to provide stable estimates of test subjects' performance by its statistical power \cite{card-etal-2020-little}. % emphasize this point and find small tests might lead to an overestimation of test results.
reporting statistical significance is also an important practice \cite{benjamin2018redefine,dror-etal-2018-hitchhikers}.
The reliability of test scores also has to be supported by reliable annotation of gold labels.
In addition, the annotation design should consider inherent disagreements among humans \cite{pavlick-kwiatkowski-2019-inherent}.
SQuAD reports its data collection process and makes the scripts of score calculations publicly available to replicate the baseline systems and reported results.
\citet{miller2020effect} similarly collect examples for different text sources, such as newspaper articles and product reviews.
They find that the average performance largely drops in those text sources compared to the original SQuAD across a broad range of systems.
This finding would help stakeholders assess the generalizability of the observed test scores if the systems need to be tested in terms of general reading comprehension beyond that for Wikipedia articles.
Furthermore, the study on SQuAD reported human performance using gold answers by three annotators to ensure reliability of the test scores, but its statistical variation is not calculated.

\paragraph{4. Explanation}
This inference determines if expected scores in the test can be associated with a construct (a conceptual tool used to facilitate understanding of human behavior in psychology) %\footnote{\url{https://www.britannica.com/science/construct}}
of the target abilities.
For instance, systems that exhibit intended behavior despite varying input enable stakeholders to validate the association between their behavior and the target abilities.
For this purpose, the test must have examples that can validate various behaviors to cover the range of abilities that the test is intended to assess.
% This requirement is emphasized by \citet{borsboom2004concept} who focus on the importance of causality between observed scores and target attributes as construct validity.
References including challenge sets \cite{mccoy-etal-2019-right}, stress tests \cite{naik-etal-2018-stress}, and contrast sets \cite{gardner-etal-2020-evaluating} aim to investigate the decision boundary of the systems.  % adversarial examples \cite{wallace-etal-2019-trick}, isabelle-etal-2017-challenge, 
Avoiding dataset biases and shortcut solutions is also important for testing intended abilities \cite{gururangan-etal-2018-annotation,geirhos2020shortcut,malaviya-etal-2022-cascading}.
For SQuAD, \citet{jia-liang-2017-adversarial} find that injecting a manually crafted distracting sentence into the passage causes the systems to predict incorrectly.
Such unintended behaviors show that examples in SQuAD may be insufficient to cover a range of inputs necessary for performing the explanation inference.
% The SQuAD consists of examples naturally collected by crowdsourcing on Wikipedia passages with no modifications.
% If it was proposed with a clear intention to evaluate the abilities of question answering only on those examples, the adversarial examples with distracting passages would be out of scope in the test.
% Therefore, to discourage overclaiming and underclaiming about the interpretation of systems' performance, there should be mentioned a limitation of the dataset in terms of text source, collection process, and targeted behavior.
% In addition, the difficulty of examples has to be considered in the test, which can also be modeled by IRT. % the IRT analysis introduced in the evaluation inference.
In most NLU tasks, examples are assumed to be easily solvable by humans.
This assumption ensures the quality of examples but lacks any control over their difficulty.
However, simply collecting examples that are difficult for systems, as in adversarial data collection \cite{kiela-etal-2021-dynabench} or dataset cartography \cite{liu-etal-2022-wanli}, may be misleading because we need to control how the difficulty contributes to the target abilities in the test \cite{bowman-dahl-2021-will}. % and trace it in varying characteristics of input examples  % \cite{ettinger-etal-2017-towards}

% The next extrapolation and utilization inferences are external criteria rather than the others concerning the test content itself.
% These inferences are pertinent to the comparison of the result of the target test with that of other tests, and to the deployment the systems tested.

\paragraph{5. Extrapolation}
This inference checks if the system performance on the test successfully correlates with other criteria in related tasks and datasets.
For example, when we have a successful system in SQuAD, the system should ideally be able to perform well in a similar dataset such as Natural Questions \cite{kwiatkowski-etal-2019-natural}.
\citet{talmor-berant-2019-multiqa} and \citet{khashabi-etal-2020-unifiedqa} analyze the generalization and transfer of performance across multiple QA datasets.
A compilation of tasks provided in the same format or platform is also helpful for users to compare the performance of systems on different tasks \cite{wang2019superglue,srivastava2022beyond,liang2022holistic}.
A successful system in machine reading comprehension is also expected to pass relevant basic tasks such as semantic role labeling and named entity recognition.
Moreover, testing the system on such auxiliary tasks contributes to its sanity checks \cite{ribeiro-etal-2020-beyond}.
% Cross-lingual applications would also be worth pursuing \cite{conneau-etal-2018-xnli,artetxe-etal-2020-cross}.
These auxiliary-task practices are optional compared to other inferences because a single test set cannot realistically provide a set of different tasks as well.
Nevertheless, referring to existing datasets and aligning the task format with that of those datasets is beneficial.
% Although few tasks are proposed for related task-based analysis for SQuAD, \citet{artetxe-etal-2020-cross} propose a dataset consisting of question-answering pairs in ten languages generated by translation.
% Another direction is moving static benchmarks towards real-world language use, for which \cite{zellers-etal-2021-turingadvice} propose a task of giving advise
% This inference may look at other practical criteria such as inference speed and memory consumption of a system to ensure the reproducibility of the system's performance.

\paragraph{6. Utilization}
This inference focuses on the utility of the test results and potential social influence of the test.
The current convention in machine learning is to train and test models on static datasets.
However, this approach does not ensure that the models are properly and fairly deployable in different configurations or real-world applications. % the usefulness and availability of 
For example, \citet{ethayarajh-jurafsky-2020-utility} and \citet{ma-etal-2021-dynaboard} advocate the reporting of model statistics, such as the size, energy efficiency, and inference latency, on leaderboards to enable more informative comparison of models in terms of utility.
% \citet{bender2021dangers} also discuss the financial and environmental risks of using large-scale pretrained language models.
% \citet{de2020towards} review the ecological validity of language user interfaces. % , pointing out that the discrepancy between current tasks and realistic situations becomes critical when systems are deployed to actual interfaces that would linguistically aid users in conducting intelligent tasks.
Regarding social influence, the test should not motivate the development of systems that have harmful social biases, such as stereotypes \cite{blodgett-etal-2021-stereotyping} and gender biases \cite{sun-etal-2019-mitigating}. %, while clarifying what kinds of representations are considered to be harmful in the test and system use \cite{blodgett-etal-2020-language}. % propose practices to conceptualize and analyze biases in NLP systems.
In addition, depending on the potential applications, we have to monitor the vulnerability of systems towards adversarial attacks \cite{wallace-etal-2019-trick}. % alzantot-etal-2018-generating,
% Unfortunately, 
Several methods have been proposed to improve the robustness of systems against adversarial inputs, such as training with diverse data \cite{tu-etal-2020-empirical} and self-debiasing framework \cite{utama-etal-2020-towards}.
% To clarify the model behavior, explanation methods would also be helpful \cite[e.g.,][]{deyoung-etal-2020-eraser}.
Although SQuAD does not address these problems, subsequent studies have addressed social biases in QA \cite{parrish-etal-2022-bbq} and the applicability of adversarial examples \cite{wallace-etal-2019-universal}.  % question-answering

% To summarize, Appendix~\ref{app:checklist-practices} provides a checklist with 16 questions for users.
To summarize, our checklist questions help users find useful practices to collect evidence that validates the test interpretations.
Although ensuring that a single test set conforms to all criteria may be impractical (e.g., there may be trade-offs between the coverage and diversity of test examples with their reliability and discriminability), knowing what evidence is missing is helpful to assess the validity of the intended interpretations and develop necessary practices.

\section{Related Work}

The definition of language understanding has been discussed in various NLP tasks, including symbol grounding \cite{merrill-etal-2021-provable} and reading comprehension \cite{dunietz-etal-2020-test}.
Our work is similar to \citet{bommasani-etal-2021-opportunities} in that we do not provide concrete definitions and highlight an epistemological perspective.

%Regarding the concept of validity, 
\citet{messick1995validity} introduce six aspects to improve the validity of interpreting results in psychological measurements.
Although \citet{sugawara-etal-2021-benchmarking} associate these aspects with the requirements of designing NLU datasets, actionable practices are not proposed for these aspects. % are proposed as a catalog
Similarly, \citet{raji-etal-2021-ai} discuss the construct validity in benchmarking AI but do not aim to improve evaluation methods. % introduce a guideline for improving evaluation methods.
% In contrast, 
Our validity argument provides a step-by-step guideline for test developers to follow.  % in our study

The concept of researcher degrees of freedom is originally introduced in psychology as a pertinent factor in severe issues such as HARKing (hypothesizing after the results are known) and publication bias \cite{munafo2017manifesto}.  % , p-hacking, failure to control for bias,  % (hypothesizing after the results are known)
In terms of resolving this problem in NLP, preregistration is a potentially promising direction for research \cite{van-miltenburg-etal-2021-preregistering}.
However, it is not suited to all areas of NLU research because certain explanatory and analytic studies do not begin with a clear hypothesis.
Nonetheless, clearly stating a research goal, problem definition, data collection method, system statistics, intended use, and potential risks could be the first step towards making credible claims on the capabilities of NLU systems.

\section{Conclusions}

% Large-scale pretrained language models exhibit near-human performance in various NLP tasks, but unjustified claims about those achievements have provided the opportunity to reconsider how we define and evaluate NLU.
The prevalence of exaggerated claims about the achievements of foundation models motivates us to reconsider how we define and evaluate NLU.
% Although the Octopus Test may not be suitable for defining language understanding, its most important contribution is that it has introduced discussions on language understanding to the NLP community.
Our formulation of NLU using the response-dependent interpretation mitigates the issues of the Turing and Octopus tests; it stipulates that observers and target conditions, including tasks and abilities, must be specified.
% A test does not provide a definition but inductive evidence for stakeholders.
However, current practices for creating NLU datasets are yet to be aligned, which provides researchers with the freedom to choose convenient strategies.
To organize essential practices using a standard guideline, we introduce the validity argument, % used in psychological and educational measurements to validate the score interpretation and use.
% The argument
which guides stakeholders to collect and interpret evidence for validating that the test subject executes its intended behavior.
% We show that 
Our proposed checklist helps researchers find relevant practices for benchmarking NLU, but we continually revise it by investigating potential refutation to promote more credible NLU studies.  % reproducible

\section*{Limitations}

% We aim to facilitate scientific communication in the community by rethinking the definition and test of NLU.
Although our discussion should be applicable to all NLP tasks, it is mainly pertinent to intellectual linguistic tasks (e.g., natural language inference, reading comprehension, and commonsense reasoning) % in a limited sense) % ; or more broadly, language grounding, dialogue systems, and machine translation)
that may involve language understanding in some sense.
The main reason for this limited applicability is that such intellectual tasks are relatively more response-dependent than basic tasks (e.g., syntactic parsing and semantic role labeling) and necessitate well-designed datasets and better evaluation methods.

Our formulation of testing language understanding in Section~\ref{sec:reframing} may be theoretically incomplete and require further discussion with reference to all related fields, including philosophy, psychology, cognitive science, and artificial intelligence.
In particular, our formulation only provides an epistemological viewpoint and thus does not provide a concrete definition of language understanding % \cite[e.g., metaphysical characterization by][Section 2.6.3]{bommasani-etal-2021-opportunities} 
for avoiding confusion, which the community needs to discuss further. % \cite[e.g.,][]{merrill-etal-2021-provable}.

In Section~\ref{sec:validity-argument}, we introduce a framework that deals with current and future practices for better NLU studies.
However, the proposed static checklist should be continually revised and improved by incorporating future findings that reveal potential flaws in our methodology to construct an effective set of checklist questions.
Furthermore, although our ultimate aim is to create a language-agnostic formulation and checklist that do not depend on specific languages, we have mainly focused on studies that deal with English texts.

Our checklist is designed for NLU and not for other NLP tasks, but it can be modified and extended to other tasks such as machine translation, language grounding, and natural language generation while referring to comprehensive meta-analysis and survey studies \cite[e.g.,][]{marie-etal-2021-scientific,chandu-etal-2021-grounding,clark-etal-2021-thats}.

\section*{Acknowledgments}

We would like to thank the anonymous reviewers for their insightful and constructive feedback.
This work was supported by JSPS KAKENHI Grant Number 22K17954, JST PRESTO Grant Number JPMJPR20C4, and JST Moonshot R\&D Program Grant Number JPMJMS2011.

\bibliography{anthology,custom}
\bibliographystyle{acl_natbib}

\appendix

\section{Brief Note on Symbol Grounding} % and Understanding}
\label{app:symbol-grounding}

According to the observations in Section~\ref{sec:octopus-test-issues}, the Octopus test argument cannot support the claim that a system that learns the form alone cannot understand language (i.e., it cannot pass the Octopus test).
However, the test remains pertinent because it continues to stimulate the intuition that language models are not symbol-grounded and are therefore unlikely to understand language.

The crucial question is whether this intuition is true; we have two reasons to doubt it.
First, we may argue that the corpora used for training current NLU systems do not comprise mere forms.
A corpus comprises linguistic expressions that are produced by social interactions among humans who understand the language.
Thus, the linguistic expressions in a corpus cannot be regarded as mere physical objects that lack meaning because they have already been assigned meaning.
Second, the meaning defined by \citet{bender-koller-2020-climbing} may not agree with our understanding of the language.
For example, a layperson has only vague ideas about the satisfaction conditions of words such as ``bacteria'' or ``nicotine'' \cite[cf.][]{evans1973causal}.
This is the case for many other general terms.
Nevertheless, people manage to use these words in their lives, and it is narrow-minded to conclude that they do not actually understand the meanings of these words. 

\section{Design and Terminology of Validity Argument}
\label{app:validity-argument}

Regarding the definition of inferences, \citet{kane2006validation,kane-2012-validating} originally define four inferences (scoring, generalization, extrapolation, and implication).
However, we adopt the extended definition of \citet{chapelle2008test} with six inferences to identify clear and detailed evidence that we consider adequate for testing NLU.
We refer readers to \citet{cook2015contemporary} for a concise practical introduction of the validity argument and to \citet{chapelle2008test} for an application example of human language testing.

We briefly mention the terminology.
As the validity argument is initially proposed for psychological and educational measurements, several terms have different meanings from those in NLP and machine learning.
For example, \textit{inference} is performed by stakeholders (e.g., researchers, model developers, test creators, and users) involved in the test, whereas it is usually performed by a model to make a prediction in NLP.
\textit{Domain} in the validity argument includes any condition that specifies target behavior and experimental settings.
In contrast, it only indicates data sources or text genres in NLP.
\textit{Generalization} in the validity argument is concerned with how the experimental results are generalized to other experimental settings, similar to the reproducibility of findings. 
Conversely, it mainly refers to the property of machine learning models in NLP, which is related to the explanation inference in the validity argument (i.e., can one explain whether a model shows capabilities that are sufficiently generalizable to the target construct?).

% In NLU and AI benchmarking, \cite{}

\section{Checklist and Relevant Practices for Validity Argument}
\label{app:checklist-practices}

\begin{table*}[t!]
    \centering
    \def\arraystretch{1.2}
    \begin{tabular}{p{6em}p{14em}p{18em}} \toprule
        Inference & Checklist & Relevant Practices \\ \midrule
        Domain \newline Definition & $\square$ Does the test set describe the target tasks and domains in detail? & Taxonomy of knowledge and skills \cite{lobue-yates-2011-types,rogers-etal-2023-qa} \\
        & $\square$ Are the test items annotated with requisite knowledge and skills? & Diagnostic dataset \cite{warstadt-etal-2020-blimp-benchmark}, qualitative annotation \cite{schlegel-etal-2020-framework}   \\
        & $\square$ Are the test items reviewed by domain experts? & Data collection with experts \cite{parrish-etal-2021-putting-linguist}, contrast sets \cite{gardner-etal-2020-evaluating} \\
        Evaluation & $\square$ Does the test set have structured metrics for its subtasks? & Sub-questions as semi-structured explanation \cite{wolfson-etal-2020-break,geva-etal-2021-aristotle} \\
        & $\square$ Are the test items difficult and sufficiently discriminative to compare systems? & Item response theory \cite{rodriguez-etal-2021-evaluation,vania-etal-2021-comparing}, dataset cartography \cite{swayamdipta-etal-2020-dataset}, crowdsourcing protocol design \cite{nangia-etal-2021-ingredients} \\
        Generalization & $\square$ Does the test set describe the data collection process and baseline systems for reproducibility? & Templates of dataset specification \cite{bender-friedman-2018-data,gebru2021datasheets}, reproducibility checklist \cite{pineau2021improving} \\
        & $\square$ Do the test items address annotation disagreement? & Taxonomy of disagreement \cite{jiang-de-marneffe-2022-investigating}, modeling annotation distribution \cite{chen-etal-2020-uncertain,nie-etal-2020-learn} \\
        & $\square$ Does the test set produce a statistically significant result? & Statistical test \cite{dror-etal-2018-hitchhikers,sadeqi-azer-etal-2020-claims}, statistical power \cite{card-etal-2020-little}, instability analysis \cite{zhou-etal-2020-curse} \\
        Explanation & $\square$ Is the test set designed to mitigate shortcut solutions? & Input ablation \cite{gururangan-etal-2018-annotation}, competency problems \cite{gardner-etal-2021-competency},  adversarial filtering \cite{zellers-etal-2019-hellaswag}  \\
        & $\square$ If the training set is provided, does the test set contain out-of-distribution examples? & Diagnosis of heuristics \cite{mccoy-etal-2019-right}, stress tests \cite{naik-etal-2018-stress,saha-etal-2020-conjnli}, contrast sets \cite{gardner-etal-2020-evaluating} \\
        & $\square$ Is there any control over the difficulty of examples? & Simplified auxiliary questions \cite{sutcliffe2013QA4MRE}, human--machine collaboration \cite{bartolo-etal-2022-models,liu-etal-2022-wanli} \\ %  test subsets with different difficulty \cite{clark-etal-2018-think,berzak-etal-2020-starc} \\
        Extrapolation & $\square$ Does the test set encourage performance comparison using related tasks and datasets? & Cross-dataset generalization analysis \cite{talmor-berant-2019-multiqa}, compilation of tasks \cite{wang2019superglue,srivastava2022beyond} \\
        & $\square$ Does the test include basic tasks for sanity check? & Checklist of basic tests for task-relevant linguistic capabilities \cite{ribeiro-etal-2020-beyond} \\
        Utilization & $\square$ Does the test set support the utility enhancement of systems? & Reporting practical statistics \cite{ethayarajh-jurafsky-2020-utility,ma-etal-2021-dynaboard} \\
        & $\square$ Does the test set include disincentive for embedding harmful social biases? & Underspecified questions \cite{li-etal-2020-unqovering}, quantifying representational harms \cite{mehrabi-etal-2021-lawyers}, bias types \cite{blodgett-etal-2020-language} \\
        & $\square$ Does the test set promote the mitigation of the system vulnerability against adversarial attacks? & Universal adversarial triggers \cite{wallace-etal-2019-universal}, data augmentation \cite{min-etal-2020-syntactic}, self-debiasing framework \cite{utama-etal-2020-towards} \\ % are there good alternatives for "encourage"?
        \bottomrule 
    \end{tabular}
    \caption{Overview of validity inferences, checklist questions, and relevant practices for NLU.}
    \label{tab:checklist-practice}
\end{table*}

Table~\ref{tab:checklist-practice} summarizes the checklist for collecting evidence in the validity argument and recent relevant practices that have been proposed in the NLU study.
We elaborate on the checklist questions and practices as follows:

\subsection{Domain Definition}

\paragraph{Does the test set clearly describe the target tasks and domains in detail?}
This question requires the test description for specifying what task a test aims to evaluate in what domain, rather than aiming for general language understanding \cite{raji-etal-2021-ai}.
The test description includes a general goal of the test, test format, intended task, data source, and potential application.
To describe target knowledge and skills, test developers can develop their own taxonomy or use existing taxonomies such as for linguistic phenomena \cite{warstadt-etal-2020-blimp-benchmark}, commonsense types \cite{lobue-yates-2011-types,sap-etal-2019-atomic}, science questions \cite{boratko-etal-2018-systematic}, reading comprehension \cite{sugawara-etal-2017-evaluation,dunietz-etal-2020-test}, and QA in general \cite{rogers-etal-2023-qa}.

\paragraph{Are the test items annotated using requisite knowledge and skills?}
The detailed annotation of required knowledge and skills with test items helps stakeholders interpret the strengths and weaknesses of the models tested and to associate them with relevant tasks \cite{doshi-velez-kim-2017-rigorous,schlangen-2021-targeting}.
Annotation can be performed in two main ways: create diagnostic examples with constraints such as keywords and templates \cite{rogers-etal-2020-getting,warstadt-etal-2020-blimp-benchmark} or annotate labels after collecting examples with no constraints \cite[][inter alia]{schlegel-etal-2020-framework}.
Nonetheless, the appropriate granularity of the annotation has to be decided depending on how much detail the stakeholders require to analyze the model behavior.
This consideration is essential because annotating detailed knowledge and skills unambiguously is difficult, even for experts.

\paragraph{Are the test items reviewed by domain experts?}
Ideally, the appropriateness of test items, given the target task, should be reviewed by experts who are familiar with the task.
\citet{parrish-etal-2021-putting-linguist} show that involving experts during data collection improves the quality of crowdsourced data by identifying artifacts.
This expert-in-the-loop design for collecting examples may be a promising direction to target intended linguistic phenomena.
Nevertheless, we must ensure that the distribution of data is not unintentionally biased towards a limited variety of linguistic phenomena \cite{bowman-dahl-2021-will}.
\citet{gardner-etal-2020-evaluating} asked experts who created the source dataset of a target task to modify task examples to ensure that the difference between the original and contrastive examples produced different gold labels.
Although hiring experts to craft test examples from scratch is expensive, at least reviewing (and annotating) some of the collected examples by the test developers (as experts) contributes to sourcing test examples that require their target task correctly.

\subsection{Evaluation}
\label{app:validity-argument-evaluation}

\paragraph{Does the test set have structured metrics for its subtasks?}
Because of the possibility of shortcut solutions \cite{geirhos2020shortcut} that circumvent intended solutions, merely observing the final output does not necessarily guarantee that the test subject performs the task precisely.
Given that a generated explanation about the answering process cannot be evaluated straightforwardly \cite{clark-etal-2021-thats}, asking about a (semi-)structured reasoning path may be a useful approach.
For example, several benchmarks require the completion of reasoning process in addition to the main QA task \cite{Bhagavatula2020Abductive,inoue-etal-2020-r4c,wolfson-etal-2020-break,geva-etal-2021-aristotle,saha-etal-2021-explagraphs}.

\paragraph{Are the test items sufficiently difficult and discriminative to compare systems?}
Item response theory is a standard way to characterize the difficulty and discriminability of test examples while modeling the ability of test subjects \cite{lalor-etal-2016-building,rodriguez-etal-2021-evaluation,vania-etal-2021-comparing,byrd-srivastava-2022-predicting}.
In the benchmarking of NLU models, a test needs to enable meaningful comparisons between models including the baseline.
If all test examples are exceedingly easy or difficult, or if there are many ambiguous examples, no significant difference in evaluation measures can be observed.
Item response theory helps test developers analyze test examples and control the distribution of difficulty and discriminability.

\subsection{Generalization}

\paragraph{Does the test set describe the data collection process and baseline systems for reproducibility?}
This requirement includes critical aspects to ensure the reproducibility of the study, such as data sources, how the annotators are employed, annotation procedure, annotation instructions, platform or software used for collection, quality control, experimental design, and baseline systems.
It is also beneficial to identify potential biases unintentionally embedded by annotators \cite{geva-etal-2019-modeling}, although the annotation instructions and examples need to be carefully presented to mitigate such biases \cite{parmar-etal-2023-dont}.
To describe the data collection process, using templates of dataset specifications, such as data statements \cite{bender-friedman-2018-data} and datasheet \cite{gebru2021datasheets} (especially their data collection part), is helpful.
If crowdsourcing is used in the data collection, reporting payment methods is also encouraged to guarantee ethical fairness \cite{kummerfeld-2021-quantifying,shmueli-etal-2021-beyond}.
If a paper proposing a new dataset includes baseline machine learning systems, ensuring the reproducibility of baseline systems is also important \cite{pineau2021improving}.

\paragraph{Do the test items address annotation disagreement?}
The dataset needs to address ambiguous test items to produce reliable results.
However, ambiguity may not be noise in annotation but an inherent property of examples \cite{zhang-etal-2017-ordinal,pavlick-kwiatkowski-2019-inherent}.
% Thus, the test creators do not have to completely remove ambiguous examples.
% However, they have to consider such ambiguity in the annotation scheme and task design.
Therefore, in addition to designing a careful procedure to take care of ambiguous and under-specified examples \cite[e.g.,][]{boyd-graber-borschinger-2020-question}, modeling the ambiguity itself can also be a meaningful task in NLU.
For example, several studies tackle the task of modeling the distribution of human votes for the labels in the natural language inference task \cite{chen-etal-2020-uncertain,nie-etal-2020-learn,meissner-etal-2021-embracing,zhang-de-marneffe-2021-identifying,zhou-etal-2022-distributed}. % defeasible NLI: rudinger-etal-2020-thinking
A taxonomy of disagreement \cite{jiang-de-marneffe-2022-investigating} is also a useful reference in the qualitative analysis of ambiguous cases.

\paragraph{Does the test set produce a statistically significant result?}
When comparing the performance between systems, choosing an appropriate statistical test is critical to prove that the observed performance difference is statistically significant \cite{dror-etal-2018-hitchhikers,sadeqi-azer-etal-2020-claims}.
In testing, a sufficient number of examples is necessary to detect a true effect of the performance improvement.
\citet{card-etal-2020-little} suggest that statistical power should be analyzed before performing evaluation.
Similarly, \citet{zhou-etal-2020-curse} analyze the performance instability in popular benchmarks, suggesting the reporting of decomposed variance measures and use of diverse analysis datasets.

\subsection{Explanation}

\paragraph{Is the test set designed to mitigate shortcut solutions?}
In the current standard of machine learning, most NLU datasets are based on the training, validation, and test split.
Although this study focuses on contributing to the methodological improvement of the test phase, the distribution between training and test split may affect the test results;
machine learning models are generally good at exploiting superficial patterns from training examples.
Identifying these patterns helps the models make accurate predictions for test examples; however, this approach does not work well for out-of-distribution (OOD) examples \cite{damour-etal-2020-underspecification,geirhos2020shortcut}.
For example, analysis of spurious correlations between gold labels and tokens reveals potential shortcut solutions \cite{gururangan-etal-2018-annotation,gardner-etal-2021-competency}.
However, \citet{schwartz-stanovsky-2022-limitations} note that controlling balancing methods is difficult for spurious correlations, such as data augmentation \cite{sharma-etal-2018-tackling} and adversarial filtering \cite{zellers-etal-2019-hellaswag,bras2020adversarial}, because these methods may diminish meaningful signals. 
Therefore, they suggest alternative methods such as adding rich contexts and stopping large-scale fine-tuning.
Analysis of input ablation may also be a useful practice to filter out easy examples, such as by hiding the premise in natural language inference \cite{gururangan-etal-2018-annotation} and removing question tokens in machine reading comprehension \cite{feng-etal-2018-pathologies,kaushik-lipton-2018-much,yu-etal-2020-reclor}.
\citet{malaviya-etal-2022-cascading} find that monitoring heuristic annotation strategies among crowdworkers may improve the quality of collected QA examples.

\paragraph{If the training set is provided, does the test set contain OOD examples?}
Similar to the previous question, to properly associate the target behavior with the intended skill, we have to ensure the generalizability and robustness of the models towards diverse examples in the target task.
Given the possibility of shortcut solutions, the test set has to contain OOD examples, that is, ones collected in a different manner to those used for the training examples \cite{ettinger-etal-2017-towards,linzen-2020-accelerate}.
This line of research includes adversarial examples \cite{jia-liang-2017-adversarial,glockner-etal-2018-breaking}, diagnosis sets for syntactic heuristics \cite{mccoy-etal-2019-right}, stress test evaluation \cite{naik-etal-2018-stress,saha-etal-2020-conjnli}, and contrast sets for probing decision boundaries of the models \cite{gardner-etal-2020-evaluating}, among others.
Adversarial data collection \cite{bartolo-etal-2020-beat,nie-etal-2020-adversarial,kiela-etal-2021-dynabench} can be an effective method of perturbing and expanding the distribution of collected items, but users should take extra care in collected items such that they are properly aligned to the target skills \cite{bowman-dahl-2021-will,kaushik-etal-2021-efficacy}.
\citet{wallace-etal-2022-analyzing} find that iterating rounds of adversarial data collection improves the quality of collected data.

\paragraph{Is there any control over the difficulty of items?}
Taking control over the item difficulty is vital to evaluating the proficiency of target skills.
However, it appears to be overlooked to capture the degree of skill proficiency in current NLU research.
This gap in evaluation can be attributed to the difficulty of instantiating the degree of proficiency of a target skill as examples with different difficulties.
In existing datasets, the number of reasoning steps in multi-hop QA may play this role \cite{yang-etal-2018-hotpotqa,wolfson-etal-2020-break}.
\citet{sutcliffe2013QA4MRE} provide auxiliary questions that are simplified variants of the main questions and require fewer reasoning steps than the main questions.
\citet{bartolo-etal-2022-models} and \citet{liu-etal-2022-wanli} have proposed a human--machine collaboration approach, that is, using a generator-in-the-loop data collection method for effectively helping annotators to enhance their creativity.
Although it is exceedingly coarse for skill-wise analysis, several datasets provide subsets of the test set with different difficulties \cite{clark-etal-2018-think,lai-etal-2017-race,yu-etal-2020-reclor,berzak-etal-2020-starc}.
Item response theory also contributes to characterizing the item difficulty (refer to the second question in Appendix~\ref{app:validity-argument-evaluation}), but the test creators have to use human responses for attributing the test scores to the target construct.

\subsection{Extrapolation}

\paragraph{Does the test set encourage performance comparison using related tasks and datasets?}
The study of model development in NLU usually uses multiple datasets to report the performance of proposed models.
However, stakeholders can choose datasets on which the proposed models perform well and refrain from reporting relatively lower scores on other datasets.
To prevent this unfair practice, the test set needs to indicate similar datasets for reference and suggest that the test users evaluate their models on those datasets.
Beyond a single task, using a collection of tasks in a single platform to facilitate the comparison of system performance across different tasks is informative \cite{wang2019superglue,hendrycks2021measuring,srivastava2022beyond,liang2022holistic}.
Given that the language understanding capabilities may not depend on what language humans speak, cross-lingual applications are worth pursuing \cite{conneau-etal-2018-xnli,artetxe-etal-2020-cross}.

\paragraph{Does the test set include relevant basic tasks for sanity check?}
The previous requirement is also applicable to basic tasks that are expected to be involved in the target NLU task.
\citet{ribeiro-etal-2020-beyond} have proposed a checklist approach that uses three types of auxiliary tasks: minimal functional test, invariance test, and directional expectation test.
Depending on the primary target task, composing subtasks into a checklist enables the system developers to probe their system in detail and detect unintended errors.
This approach is helpful if we can create test cases for requisite knowledge and skills that are presumed in the domain definition.

\subsection{Utilization}

\paragraph{Does the test encourage the reporting of system statistics for utility?}
Static leaderboards of benchmark datasets usually tell us which system is better than others in terms of simple evaluation metrics such as accuracy.
However, they do not tell us about which system is most useful under conditions such as computational budget and inference time.
Therefore, \citet{ethayarajh-jurafsky-2020-utility} and \citet{ma-etal-2021-dynaboard} advocate the reporting of model statistics, such as the size, energy efficiency, and inference latency.
In a similar attempt, \citet{min-etal-2020-efficientqa} propose a shared task for efficient open-domain QA models, comparing the QA performance under limited memory budgets.
\citet{bender2021dangers} discuss the financial and environmental risks of using large-scale pretrained language models.
\citet{de2020towards} review the ecological validity of language user interfaces.

\paragraph{Does the test set include disincentive for embedding harmful social biases?}
Language models and word embeddings are often found to contain harmful social biases, such as stereotypes \cite{sun-etal-2019-mitigating,blodgett-etal-2020-language,blodgett-etal-2021-stereotyping}.
To date, studies on social biases in NLU datasets have been limited \cite[e.g.,][]{rudinger-etal-2017-social}.
However, \citet{li-etal-2020-unqovering} find that underspecified questions with ambiguity in their answer candidates reveal various stereotypes.
% clarifying what kinds of representations are considered to be harmful in the test and system use \cite{blodgett-etal-2020-language}.
\citet{mehrabi-etal-2021-lawyers} propose quantifying representational harms in commonsense knowledge bases.
Although it might not be harmful, falsehood should also be mitigated in foundation models \cite{lin-etal-2022-truthfulqa}.

\paragraph{Does the test set encourage the mitigation of the system vulnerability against adversarial attacks?}
Improving the robustness of systems against OOD input is one of the main concerns in the current NLP community.
As \citet{wallace-etal-2019-universal} demonstrate, NLU system predictions are easily changed by adversarial inputs, calling for improvements in the robustness against OOD data including malicious attacks.
For this purpose, various methods applicable to NLU systems have been proposed, such as training with diverse data \cite{tu-etal-2020-empirical}, systematic data augmentation \cite{min-etal-2020-syntactic,wu-etal-2021-polyjuice}, and self-debiasing framework \cite{utama-etal-2020-towards}.
\citet{wang-etal-2022-measure} provide a broad survey of datasets and methods for measuring and improving the robustness of NLP models.

\section{Potential Arguments and Discussions}

In this section, we discuss potential arguments on how we need to deal with our proposed framework.

\paragraph{``Who should use the proposed framework and for what purpose?''} % should be used to evaluate a dataset, a general task or how people use a dataset in general''}
The framework is related to the entire experimental design including the construction of a dataset and its use for evaluating systems. Thus it should be mainly used by researchers who release datasets and propose their evaluation procedure, but dataset users (i.e., system developers in most cases) can also use this framework to see if the proposed procedure is well-designed, revise it if necessary, and validate their interpretations of system behavior.

\paragraph{``Should we address all checklist questions to construct the validity argument? It may not be always possible to create such an argument owing to several constraints such as cost and data.''}
Addressing all questions in the checklist may not be practical.
Despite the difficulty in creating thorough frameworks, the checklist contributes to clarifying the potential limitations of a study.
Researchers and developers are encouraged to make justified accurate claims about their achievements.
In addition, as the community grows and new practices are introduced, including all necessary practices in a single study is expected to become infeasible.
Nonetheless, our framework provides a comprehensive reference to collect the necessary evidence for validating NLU evaluation.

\paragraph{``Why is the discussion of the Octopus test, as well as the response-dependent formulation of language understanding, prerequisite for creating the validity argument?''}
Our response-dependent formulation highlights the difficulty of developing a concise definition for NLU.
The definition of NLU depends on the goal of stakeholders who use the test results as evidence.
We suspect that exaggeration and underestimation in NLP can be attributed to the confusion about this response-dependent property of NLU; therefore, we discuss the problems of the Octopus test while referring to those of the original Turing test. 
Our focus is language understanding; hence, discussing the Turing test alone is not sufficient.

\end{document}